\documentclass[conference]{IEEEtran}
\usepackage{times}

\usepackage[numbers]{natbib}
\usepackage{multicol}
\usepackage[bookmarks=true]{hyperref}

\usepackage{amsmath} 
\usepackage{amssymb}  

\usepackage{morefloats}
\usepackage[pdftex]{graphicx}
\graphicspath{{./images}}
\DeclareGraphicsExtensions{.pdf,.jpeg,.png,.jpg,.PNG}

\usepackage{array}
\usepackage[noend]{algorithmic}
\usepackage{algorithm}
\usepackage{setspace}

\usepackage{caption}
\usepackage{subcaption} 
\usepackage{placeins}

\begin{document}

\title{Simple Swarm Foraging Algorithm Based on Gradient Computation}

\author{\authorblockN{Simon O. Obute}
\authorblockA{School of Computing\\
University of Leeds\\
Leeds, UK, LS2 9JT\\
Email: scoo@leeds.ac.uk}
\and
\authorblockN{Mehmet R. Dogar}
\authorblockA{School of Computing\\
	University of Leeds\\
	Leeds, UK, LS2 9JT\\
	Email: M.R.Dogar@leeds.ac.uk}
\and
\authorblockN{Jordan H. Boyle}
\authorblockA{Department of Mechanical Engineering\\
University of Leeds\\
Leeds, UK, LS2 9JT\\
Email: J.H.Boyle@leeds.ac.uk}}



%

\maketitle

\begin{abstract}
	Swarm foraging is a common test case application for multi-robot systems. In this paper we present a novel algorithm for controlling swarm robots with limited communication range and storage capacity to efficiently search for and retrieve targets within an unknown environment. In our approach, robots search using random walk and adjust their turn probability based on attraction and repulsion signals they sense from other robots. We compared our algorithm with five different variations reflecting absence or presence of attractive and/or repulsive communication signals. Our results show that best performance is achieved when both signals are used by robots for communication. Furthermore, we show through hardware experiments how the communication model we used in the simulation could be realized on real robots.
\end{abstract}

\section{Introduction}\label{sec_intro}

Swarm robotics applies swarm intelligence behaviours observed in natural swarms to multi-robot applications \cite{Bayindir2016}. Swarms in nature have the impressive ability of accomplishing complex tasks through the interaction of simple agents with each other and/or their environments. For example, ants are able to forage food from distant locations, beyond the sensory capabilities of an individual swarm agent, by following pheromone trails which other ants have laid to connect food sources and the swarm nest. Natural swarms are made up of relatively simple individuals acting cooperatively to achieve swarm objectives in robust, flexible and scalable ways \cite{Brambilla2013a}. They serve as the main drive for research in swarm robotics. An individual agent in the group does not have access to global knowledge of the world and relies only on interaction with its immediate environment  (and sometimes memory of previous experience) to make autonomous control decisions.

Foraging is a canonical test case for swarm robotics applications, which involves collective search and transport of objects (or food) to a specific deposit site (or nest) \cite{Zedadra2017}. It integrates research in robotic exploration, navigation, object identification, manipulation and transport within a single relatively simple (in comparison to the task) robot platform. Foraging also has diverse real-world applications such as cleaning, harvesting, search and rescue, hazardous waste clean-up, landmine clearance and planetary exploration \cite{Winfield2009}. Swarm robot foraging, like  other swarm robotics implementations, draws inspiration from  biological parallels such as foraging approaches observed in ants and bees. A key means of achieving cooperation among swarm members during foraging is through communication. This has mostly been realized through a shared memory, communication through the environment or direct communication \cite{Bayindir2016}.

In shared memory implementations, all robots have access to a shared medium to write and read information, which gives swarm robots a global means of communication. In \citep{Arkin1993}, robots that locate attractants (objects to forage) use the shared medium to notify all swarm members of the target's location. \citet{YifanCai2014} used the global knowledge of percentage of targets found and environment covered by the swarm to adapt the foraging strategy. Major drawbacks of this approach are issues related to scalability, increased complexity of individual robots and inconsistency with the swarm paradigm of decentralized control, local sensing and communication. A closely related approach is the use of a central nest as means of exchanging information among the swarm, where only robots within a limited range of the nest are able to communicate with it \cite{Hecker2015,Lu2018}. However, this makes the central nest a single point of failure for the swarm.

\begin{figure}[t]
	\centering
	\includegraphics[width=0.5\textwidth]{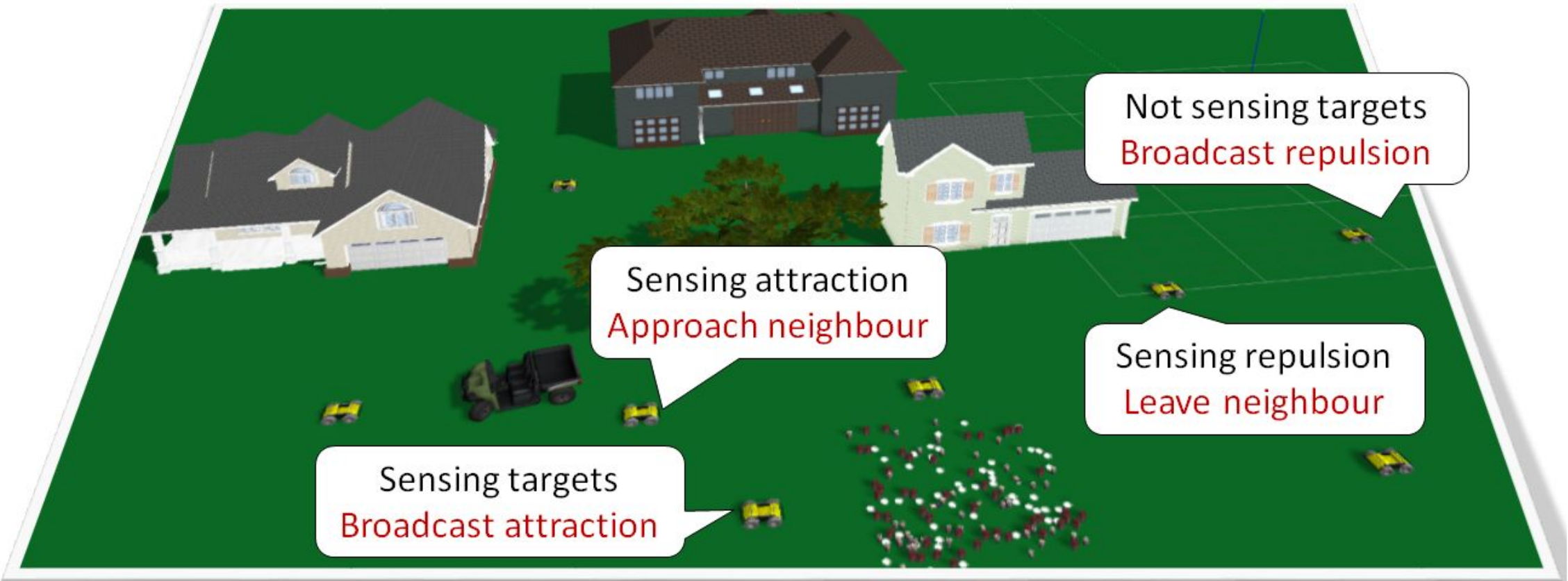}
	\caption{In direct communication, robots exchange information with neighbours based on their local environment. The neighbouring robots use this information to adapt their search/foraging strategy. In the present work, communication is limited to selectively broadcasting simple attractive and respulsive signals.}
	\label{fig_direct-communication}
\end{figure}

Achieving cooperation using the environment as a medium of communication involves modification of the search space using ``markers" or ``beacons" that provide information to guide foraging robots. The work in \cite{Hoff2012} used robots to form stationary beacon networks that broadcast hop counts of their proximity to the nest and target locations, thereby forming a gradient to be used by foraging robots to locate and transport targets to the nest. In \cite{Zedadra2015}, robots made use of the Cooperative-Color Marking Foraging Agents algorithm to mark out areas that have been covered and lay pheromone trails when returning to the nest if they successfully located a food source. When searching, robots were repulsed from explored areas and attracted to pheromone trails. The repulsive pheromone model was also used in \cite{Lima2017}. In their work, robots also kept short-term memory of visited locations to prevent returning to recently explored sites. In \cite{Ducatelle2011}, pre-deployed flying robots were used to form beacon networks by attaching to ceilings. They communicated instructions to foraging ground robots about paths connecting target locations to the nest. A major challenge for this communication model is finding an effective means of ``marking" the environment without sacrificing some robots as stationary beacons, which would affect the scalability of the swarm.

In direct communication (Fig. \ref{fig_direct-communication}), robots adapt their behaviour to improve foraging efficiency based on information exchanged with neighbouring robots. This information could be measuring a robot's position relative to their neighbours, which was used in \cite{Hoff2012} to maintain a formation of foraging robots that performed a circular sweep around the nest. Robots in \cite{Arkin1992} used their relative position to neighbours to avoid nearby robots as they searched for attractors (targets) to forage. In \cite{Rybski2008}, robots used light on top of their bodies to attract neighbouring robots when they located targets. In \cite{Dai2009}, robots communicated their foraging success to neighbouring robots they encountered in the environment. This information was used to adjust the foraging threshold of a robot, which represented its likelihood of going out to search for food whenever it returned to the nest. In \cite{Pitonakova2016}, robots mimicked the foraging behaviour of honeybees. Robots returning from scouting for targets communicated the target locations to other robots at the nest. The listening robots probabilistically decided whether to forage from the advertised food source. Robots exchanged foraging success and neurocontroller weights with other robots they encountered in \cite{Perez2017}. They then used this information to adapt their neurocontroller weights to improve foraging efficiency. Direct communication approaches overcome the scalability and complexity issues of shared memory, and the implementation challenges of environmental communication. However, there are challenges regarding what type of information should be exchanged, handling interactions with multiple neighbours simultaneously and robustness/reliability of communication media. Our approach shows that analogue sound signals can be used as a simple but effective means of direct communication among a swarm of foraging robots.

The biological foundation for our approach is the chemotaxis behaviour observed in micro-organisms such as the \textit{Escherichia coli} bacterium \cite{Nurzaman2011} and \textit{Caenorhabditis elegans} nematode \cite{Ward1973}, whose motion are characterized by near-linear runs (`swimming' mode) with occasional turns (`turning' mode) that randomize the direction of the next run. The probability that an individual \textit{E. coli} or \textit{C. elegans} will  perform a turn at any given moment depends on the temporal concentration gradient of a sensed chemical attractant (or repellent) in the environment. If conditions are improving (positive attractant gradient or negative repellant gradient) the organism reduces its turn probability, whereas worsening conditions lead to an increased probability. The simple but elegant approach of responding to the change over time means that a `single pixel' non-directional analogue sensor is sufficient. 

\begin{figure}[t]
	\centering
	\includegraphics[width=0.5\textwidth]{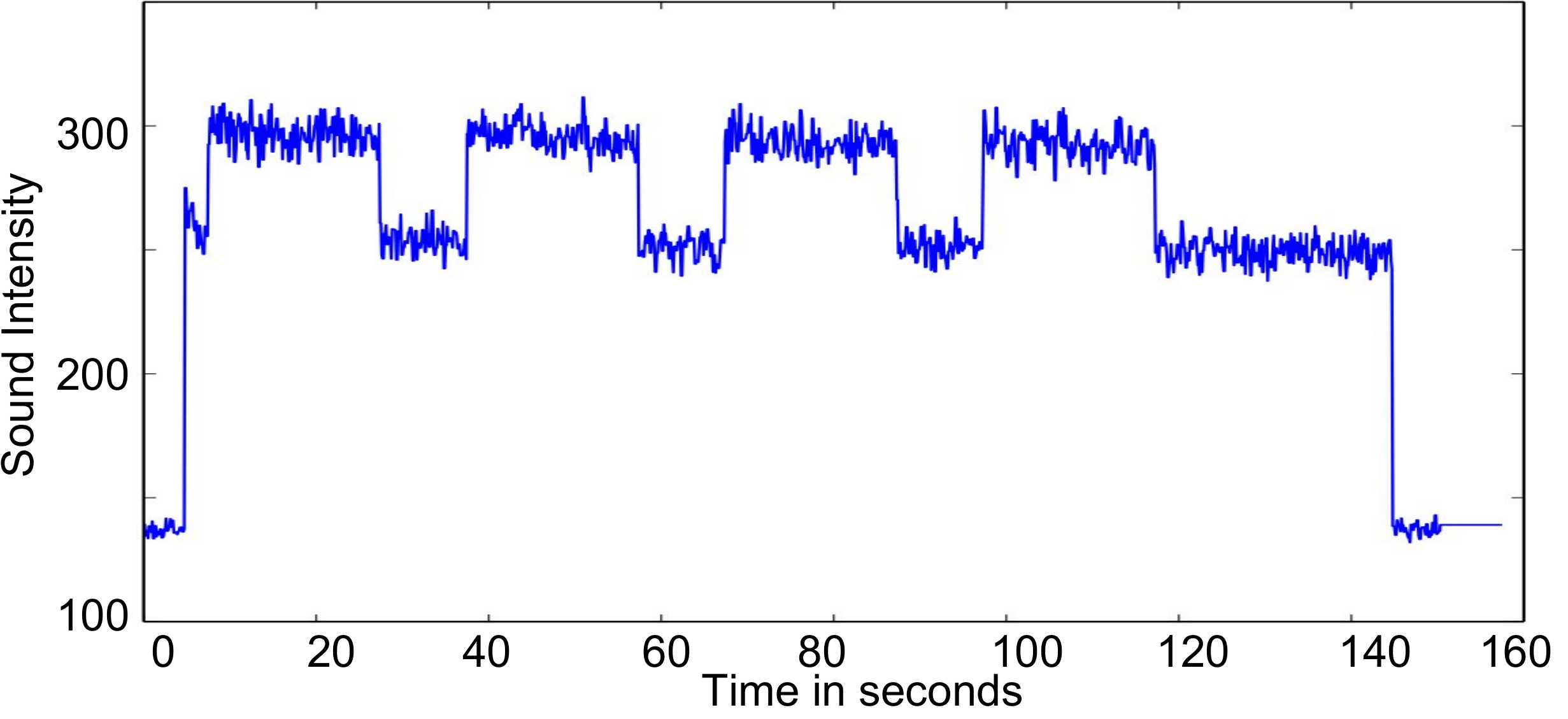}
	\caption{Simple experiment using white noise from two speakers that demonstrates increased intensity when multiple sound sources are active (appearing as intermittent peaks in the plot).}
	\label{fig_20180618083518data_stattest2}
\end{figure}

In our proposed swarm algorithm, the robots themselves serve as signal sources that create `concentration gradients' (repellent and attractant) which are used by neighbouring swarm members to improve their target search. Furthermore, the robots sense and/or broadcast these signals selectively depending on which state they are in, thereby creating a dynamic sensory landscape. These properties make our approach distinct from the biological foundations of our algorithm and its previous robotic impmlementations \cite{Nurzaman2009}. While it is primarily a direct communication method, this approach also shares some properties with environmental communication in that it exploits natural properties of the environment. By using sound as the communication signal, gradient formation and superposition of multiple signal sources happens `for free' through environmental physics (as shown in Fig. \ref{fig_20180618083518data_stattest2}).

Our approach emphasizes extreme algorithmic simplicity and demonstrates the power of minimalist bio-inspired navigation algorithms. By foregoing complex communication systems , the algorithm lends itself to simple, low cost hardware implementations. To the best of our knowledge, this is the first time this approach has been applied to swarm foraging.
 
We present this repulsion-attraction (Rep-Att) algorithm in Section \ref{sec_rep_att} and simulation results based on an idealised communication model in Section \ref{sec_analysis}. In Section \ref{sec_sound_modelling_results} we present our work on implementing the communication model in hardware platforms, and demonstrate that the algorithm still works with a realistic and noisy sound model. We also validate the basic `chemotaxis' behaviour on a real robot. Finally, concluding remarks and future directions are presented in Section \ref{sec_conclusion}.

\section{Repulsion-Attraction Algorithm (Rep-Att)}\label{sec_rep_att}
Rep-Att is based on the use of a communication mechanism whose intensity decreases smoothly with increasing distance as represented by Equation \ref{eqn_linear_intensity}, where $A^{k}_{ij}$ is signal strength of type $k$ sensed by robot $i$, located $d_{ij}$ metres away from signal source $j$ and $C_{max}$ is a robot's maximum communication range. Total signal strength sensed by a robot at a particular time, $I^{k}_{i}(t)$, at any location in the world is the sum of all signals of the same type at that location (Equation \ref{eqn_total_intensity}), where $n$ is the total number of robots and $k$ is the signal type. We consider only two signal types that robots can sense and broadcast - repulsion ($k=0$) and attraction ($k=1$) signals - within a limited range ($C_{max}$). The robots then sense gradients of individual signals ($\Delta I^{k}_{i}$) by computing the difference in signal intensity between two time steps (Equation \ref{eqn_intensity_gradient}) to use as a tool for improving swarm foraging efficiency. The influence of these signals on a robot depends on whether it is in the searching, acquiring, homing or obstacle avoidance states.

\begin{eqnarray}
	A^{k}_{ij} &=& \left\lbrace\begin{matrix}
	\dfrac{C_{max}-d_{ij}}{C_{max}} & \text{if}\quad d_{ij} \le C_{max} \\ &\\
	0,& \text{Otherwise}
	\end{matrix}\right.\label{eqn_linear_intensity}
	\\\nonumber\\
	I^{k}_{i}(t) &=& \sum_{j=1,j\neq i}^{n}A^{k}_{ij} \label{eqn_total_intensity}\\\nonumber\\
	\Delta I^{k}_{i} &=& I^{k}_{i}(t) - I^{k}_{i}(t-1) \label{eqn_intensity_gradient}
\end{eqnarray}

	\textbf{Searching State } is when a robot does not sense any target (item to be foraged) within its visual range ($V_{r}$). It broadcasts a repulsion signal ($A^{0}$) to its neighbours while using a random walk (runs of forwards motion interspersed by random turns) to search for targets. Its goal in this state is to minimize the repulsion ($I^{0}$) and maximize the attraction ($I^{1}$) signals it senses. This is achieved by detecting the change in intensity of these signals between two time steps. A robot increases its turning probability when moving in the wrong direction, i.e. when $\Delta I^{0} > 0$ or $\Delta I^{1} < 0$. Doing this increases a robot's likelihood of reorienting itself in the desired direction. On the other hand, when the robot senses a positive gradient for attraction $(\Delta I^{1} > 0)$ or a negative repulsion gradient $(\Delta I^{0} < 0)$, it reduces its turning probability, which in turn helps the robot to maintain its current direction for a longer period of time and consequently approach a region that increases its likelihood of finding a target or exploring regions with less searching robots.
	
	\textbf{Acquiring State }is when a robot detects target(s) within its visual range ($V_{r}$). It switches from searching to the acquiring state, ignores all communication signals and collects the nearest target. In cases where the robot detects multiple targets exceeding its carrying capacity, it broadcasts a range-limited attraction signal ($A^{1}$), which searching robots within communication range can sense to adjust their turning probabilities accordingly.
	
	\textbf{Homing State} is used when the robot's capacity is full. In this state, the robot heads directly towards the central deposit site (it is assumed that the direction is known) in order to deposit the collected targets. In this state, the robot ignores attraction and repulsion signals from nearby robots until it has successfully offloaded all foraged targets at the deposit site after which it switches to the searching state.
	
\begin{figure}[t]
	\centering
	\includegraphics[width=0.5\textwidth]{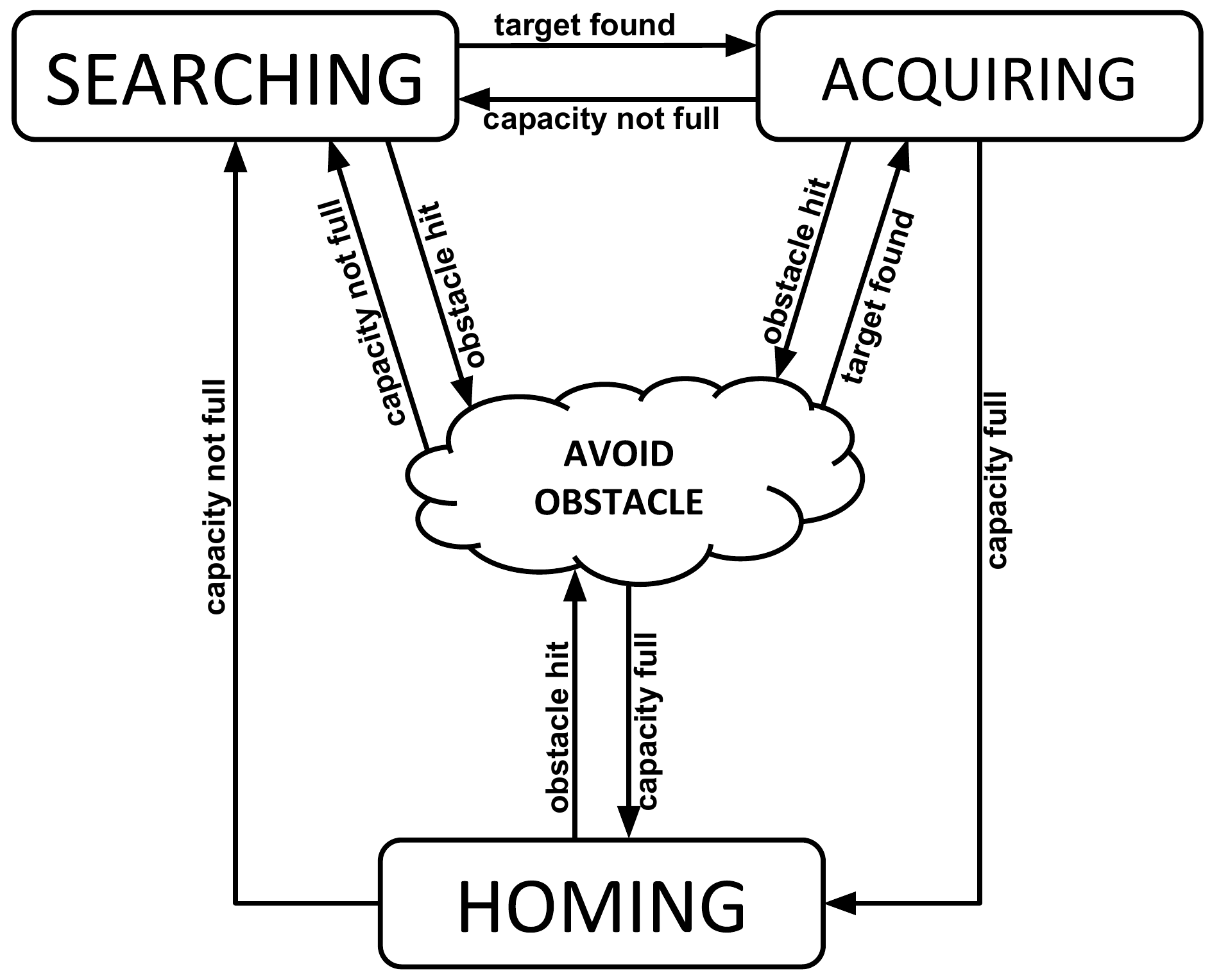}
	\caption{Finite state machine representation of foraging task, with the consideration of obstacle avoidance behaviour.}
	\label{fig_FSM-Swarm-Foraging}
\end{figure}

	\textbf{Obstacle Avoidance State }is used by robots to avoid static (nest and walls) and dynamic (other robots) obstacles. When a robot bumps into an obstacle, it avoids it by turning away from the obstacle and making a random linear motion between 0 and 1m. The robot then switches to either the searching, acquiring or homing state, depending on its local environment and availability of space to carry more targets.
The interaction between robot states is shown in Fig. \ref{fig_FSM-Swarm-Foraging}, while Algorithm \ref{alg_repAtt} illustrates the swarm foraging algorithm described in this section in psuedo code.

\begin{algorithm}[t]
{
	\begin{spacing}{1.1}
		\begin{algorithmic}[1]
			\STATE Initialize Algorithm
			\WHILE{\TRUE}
				\IF{capacity is full}
					\STATE Go home and drop collected targets
				\ELSE
					\STATE $P_{t} = P_{b}$
					\IF{$repel$}
						\IF{$\text{targets seen} == 0$ \OR $repeller$}
							\STATE Broadcast Repulsion $A^{0}_{i}$
						\ENDIF
					
						\IF{$\Delta I^{0}_{i} < 0$}
							\STATE $P_{t} = P_{b} \div G_{r}$
						\ELSIF{$\Delta I^{0}_{i} > 0$}
							\STATE $P_{t} = P_{b} \times G_{r}$
						\ENDIF
					
					\ENDIF
					\IF{$attract$}
						\IF {$\text{targets seen} > c_{m} $}
							\STATE Broadcast Attraction $A^{1}_{i}$
						\ENDIF
					
						\IF{$\Delta I^{1}_{i} < 0$}
							\STATE $P_{t} = P_{b} \times G_{a}$
						\ELSIF{$\Delta I^{1}_{i} > 0$}
							\STATE $P_{t} = P_{b} \div G_{a}$
						\ENDIF
					
					\ENDIF

					\IF{$\text{targets seen} > 0$}
					
						\STATE Go and pick up closest target
					\ELSE
					
						\STATE Random walk using $P_{t}$ as probability of turn
					\ENDIF
				\ENDIF
			\ENDWHILE
		\end{algorithmic}
	\end{spacing}
}
\caption{Swarm Foraging Algorithm}
\label{alg_repAtt}
\end{algorithm}

{
\renewcommand{\labelenumi}{\Roman{enumi}.}
From Algorithm \ref{alg_repAtt}, five  variants can be derived based on the values of the $repel$, $attract$ and $repeller$ Boolean variables:
\begin{enumerate}
	
	\item \textbf{Random Walk:}
	In this algorithm, robots do not listen to repulsion and attraction signals. They search with a constant turning probability and view other swarm robots as obstacles upon colliding with them. Thus, $repel =  false$, $attract = false$  and $repeller = false$.
	
	\item \textbf{Repeller:}
	This is based on the functionality of the algorithm presented in \cite{Arkin1992}, where robots that cannot see targets are constantly repelled by their neighbours (regardless of what the neighbours sense). To achieve this behaviour from Algorithm \ref{alg_repAtt}, $repel$ and $repeller$ are set to $true$ while attract is set to $false$.
	
	\item \textbf{Selective Repulsion:}
	Here, robots communicate repulsion signals when in the searching state and refrain from communication in other states, which is different from Repeller algorithm that repels in all states. The $repel$ variable is $true$, while $attract$ and $repeller$ are both $false$.
	
	\item \textbf{Selective Attraction:}
	For this variant, $repel$ and $repeller$ are set to $false$ while $attract$ is set to $true$. This means robots do not repel each other, but attract neighbours when they locate more targets than they can carry.
	
	\item \textbf{Rep-Att:} This is the full version of our algorithm, in which the selective attraction and selective repulsion communication mechanisms are active. $repel =  true$, $attract = true$  and $repeller = false$. It has been described in detail at the beginning of this section.
	
\end{enumerate}
}

\section{Simulation and Results (Idealised Communication)}\label{sec_analysis}
\subsection{Simulation Setup}
The Gazebo Simulator \cite{Tan2013}, with maximum step size set to 0.025 second, was used for the simulation experiments to compare the effects of presence/absence of repulsion and/or attraction signals as described in Section \ref{sec_rep_att}.

To thoroughly investigate the performance of the algorithms, five different distributions were used to initialize locations of 200 targets within a 50m x 50m world. The target distributions are one, two or four clusters, a uniform distribution, and a uniform distribution overlayed with a single cluster (HalfCluster in the figures). The goal in each world scenario is for 36 robots having limited carrying capacity and sensing range to locate the targets and return at least 90\% of them to the central deposit site (nest). Each robot in the swarm has a velocity of 0.605 m/s and spends 5 seconds processing each target it touches (i.e. to remove a target from the ground). Table \ref{table_parameters} outlines other parameters for the simulation experiments.

\begin{table}[t]
	\centering
	\caption{Parameters used in simulation and robot experiments. The same values were used for all algorithms (where the algorithm makes use of the parameter) except stated otherwise in the text.}
	\label{table_parameters}
	\setlength\tabcolsep{6pt} 
	{\renewcommand{\arraystretch}{1.1}\begin{tabular}{@{}
				>{\raggedright}p{0.25\textwidth}
				>{\raggedright}p{0.05\textwidth}
				>{\raggedright\arraybackslash}p{0.1\textwidth}@{}}\hline
			\textbf{Parameter}  & \textbf{Symbol} & \textbf{Value} \\ \hline
			Base turning probability per time step        & $P_{b}$ & 0.0025\\
			Attraction Gain    & $G_{a}$ & 10 \\
			Repulsion Gain      & $G_{r}$ & 10 \\
			Robot targets-carrying capacity   & $c_{m}$ & 5\\
			Maximum communication range           & $C_{r}$ & 15m \\
			Target detection distance             & $V_{r}$ & 3m \\
			Random turn distribution & &$\mathcal{N}(180^{0},90^{0})$
			\\\hline
		\end{tabular}
	}
\end{table}

\subsection{Results}\label{sec_results}

\begin{figure*}[!h]
	\centering
	
	\begin{subfigure}[b]{0.3\textwidth}
		\centering
		\scalebox{1}[1]{\includegraphics[width=\textwidth]{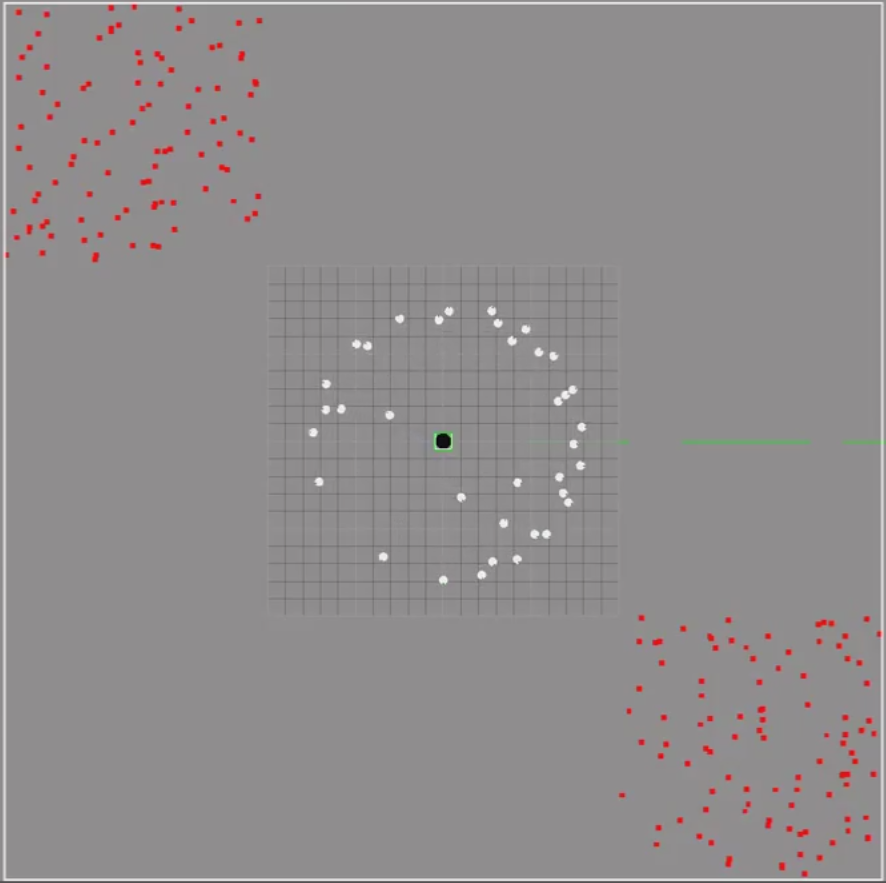}}
		\subcaption{t = 10s}
		\label{fig_t10_TwoClusterRep-Att}
	\end{subfigure}
	\hfill
	\begin{subfigure}[b]{0.3\textwidth}
		\centering
		\scalebox{1}[1]{\includegraphics[width=\textwidth]{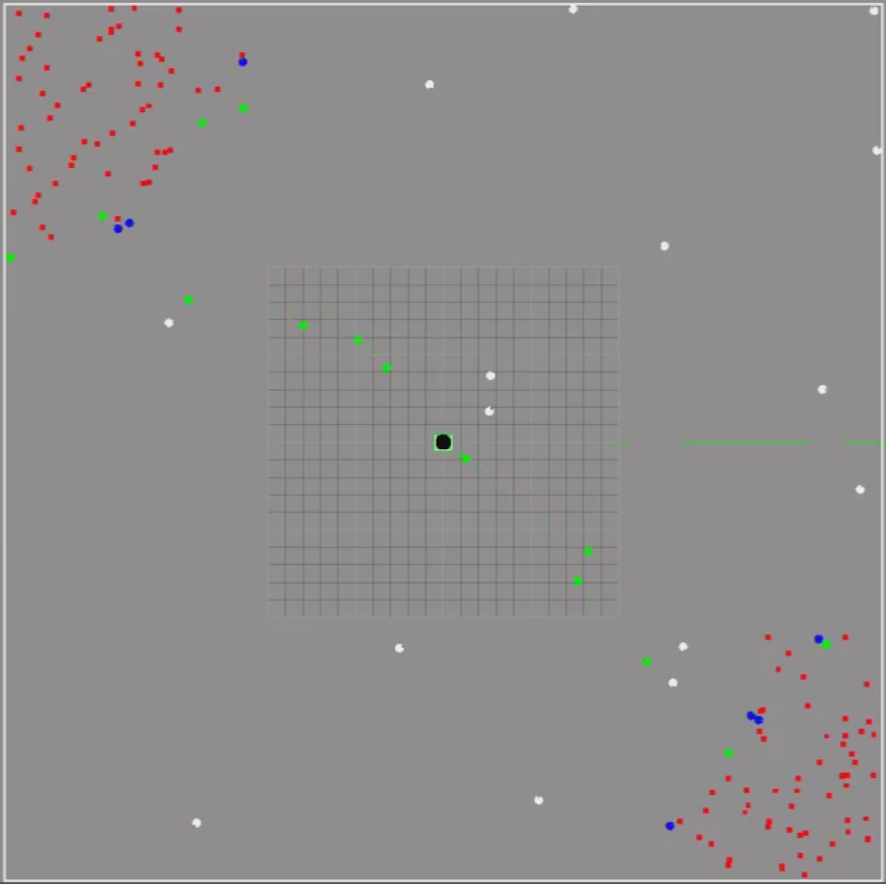}}
		\subcaption{t = 90s}
		\label{fig_t90_TwoClusterRep-Att}
	\end{subfigure}\hfill
	\begin{subfigure}[b]{0.3\textwidth}
		\centering
		\scalebox{1}[1]{\includegraphics[width=\textwidth]{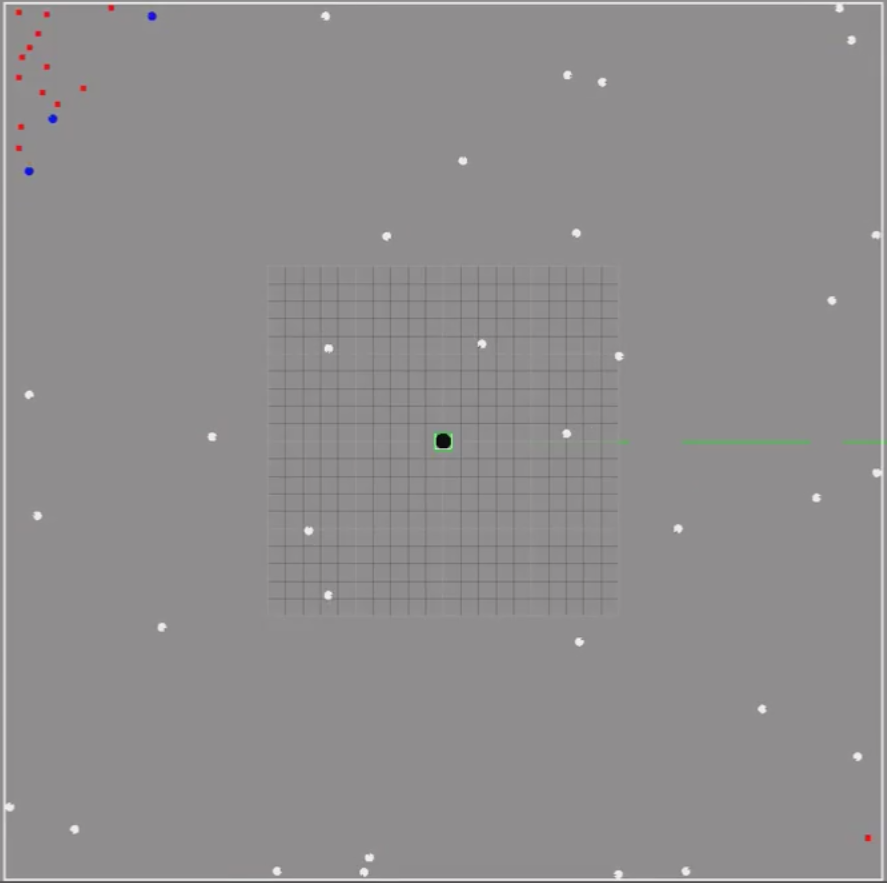}}
		\subcaption{t = 390s}
		\label{fig_t390_TwoClusterRep-Att}
	\end{subfigure}
	\caption{Simulation snapshots of Rep-Att when foraging in Two Clusters target distribution. Targets are represented in red, searching, acquiring and homing robots are white, blue and green respectively, while central nest is black. (Best viewed in colour)}
	\label{fig_snapshot_two_cluster_simulations}
\end{figure*}

Simulation snapshots for Rep-Att algorithm for the Two Clusters target distribution are shown in Fig. \ref{fig_snapshot_two_cluster_simulations} (supplementary videos are also provided). Here, we see excellent collaboration among swarm members when being controlled with the Rep-Att algorithm. Repulsion signals at the start of the simulation causes the robots to quickly leave the nest in search of targets. When robots find target clusters, they activate attraction signal to invite searching robots to cluster regions for increased exploitation of that area.

\begin{figure*}[t]
	\centering
	\includegraphics[width=0.8\textwidth]{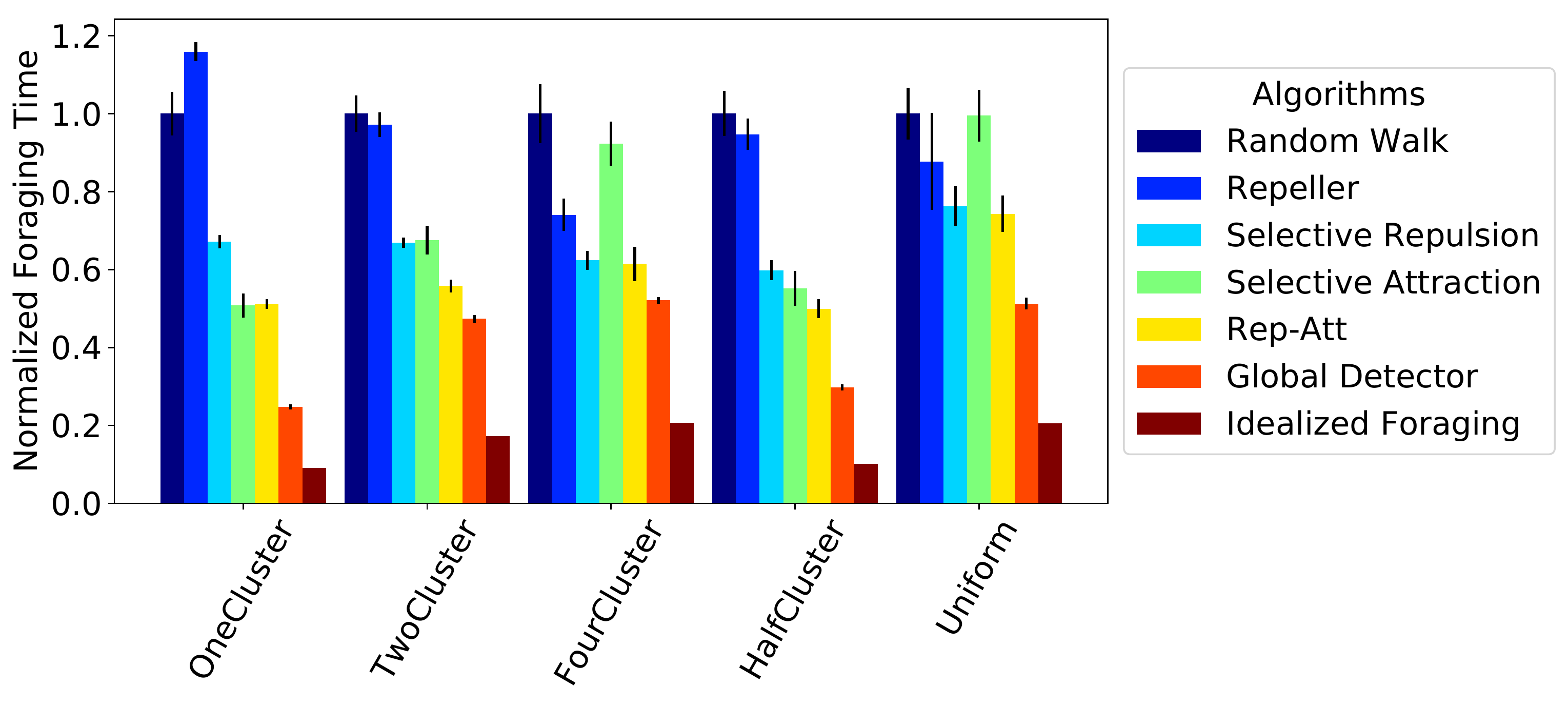}
	\caption{Normalized time taken to return 90\% of targets to nest in 5 different world scenarios. Each bar represents the mean of 30 simulation repetitions. The error bars represent 95\% confidence interval. Random walk foraging time was used as baseline for computing normalized times for the respective target distributions}
	\label{fig_limited_capacity_bar_chart}
\end{figure*}

The simulation results for the five target distributions are summarized in Fig. \ref{fig_limited_capacity_bar_chart} as the average time (normalized with the time taken by Random Walk for that distribution) taken by the swarm to successfully return 90\% of targets to the nest. Each experiment was repeated 30 times. We have also included results for two unrealistic, but superior, foraging algorithms for benchmarking purposes: Global Detector and Idealized Foraging. Global Detector is when robots in the swarm  have infinite target detection range i.e. $V_{r} = \infty$, effectively eliminating the searching state described in Section \ref{sec_rep_att}. In Idealized Foraging, we apply a statistical approach to estimate the minimum possible collection time $T_{min}$  assuming perfect task allocation and ignoring robot collisions, according to Equation \ref{eqn_idealized_detector}
\begin{align}
	\begin{split}T_{min} =& \frac{1}{N_{R}}\times \Bigg[\frac{1}{V_{R}}\times
				\bigg(\frac{N_{T}}{R_{C}} \times 2\overline{D_{TB}}\   + \\
				&\left(R_{C} - 1\right) \times \overline{D_{TN}} \bigg)
				 + 
				N_{T} \times T_{P}\Bigg]\end{split} \label{eqn_idealized_detector}
\end{align}
Where $N_{R}$ = swarm size, $V_{R}$ = robot velocity, $N_{T}$ = number of targets, $R_{C}$ = robot capacity, $\overline{D_{TB}}$ = mean distance of all targets from nest, $\overline{D_{TN}}$ = mean distance between a target and nearest $R_{C} - 1$ targets, and $T_{P}$ = target processing time.

The bar plot of Fig. \ref{fig_limited_capacity_bar_chart} shows that Rep-Att algorithm outperforms both Random Walk and Repeller algorithms on all world scenarios, giving statistically significant lower foraging times in all cases. T-test analysis conducted between Rep-Att and these two algorithms yielded a maximum p-value of 0.028 on the uniform target distribution world, while other world scenarios gave lower p-values. This shows that our Rep-Att algorithm is able to significantly improve foraging efficiency by simple adaptation of robot turning probability based on repulsion and attraction gradients sensed by individual robots.

Fig. \ref{fig_limited_capacity_bar_chart} also investigates the individual effects of attraction and repulsion signals under the different target distributions. What we see is that repulsion is most useful for the swarm when targets are less clustered in the environment. This is most evident in the four clusters and uniform targets distributions. Attraction on the other hand is most effective when resources/targets are densely clustered within few regions in the world as is evident in the one cluster case. Our Rep-Att algorithm is, therefore, able to forage efficiently across all the simulated target distributions because it integrates the contrasting strengths of repulsion and attraction within a single algorithm. 

In comparison to the unrealistic Global Detector and Idealized Foraging algorithms, Rep-Att, unsurprisingly, performs less well. However, it is interesting to note that Rep-Att is somewhat competitive when compared to the Global Detector in the Two Clusters and Four Clusters target distributions, where Rep-Att only took 17.77\% and 17.96\% more time respectively. This shows that when an environment has multiple clustered targets at various locations, repulsive signals will quickly drive multiple robots to locate target clusters. Other robots are then more likely to come within communication range of one or more attracting robot(s), thereby reducing the time they would have spent searching for targets. The difference in performance between Rep-Att and the unrealistic algorithms is more pronounced in other world scenarios because: robots spend more time searching for targets (One Cluster and HalfCluster distributions case); and robots did not sense enough targets to attract neighbours to their foraging locations (Uniform distribution case).

\begin{figure*}[h]
	\begin{subfigure}[b]{0.19\textwidth}
		\centering
		\includegraphics[width=\textwidth]{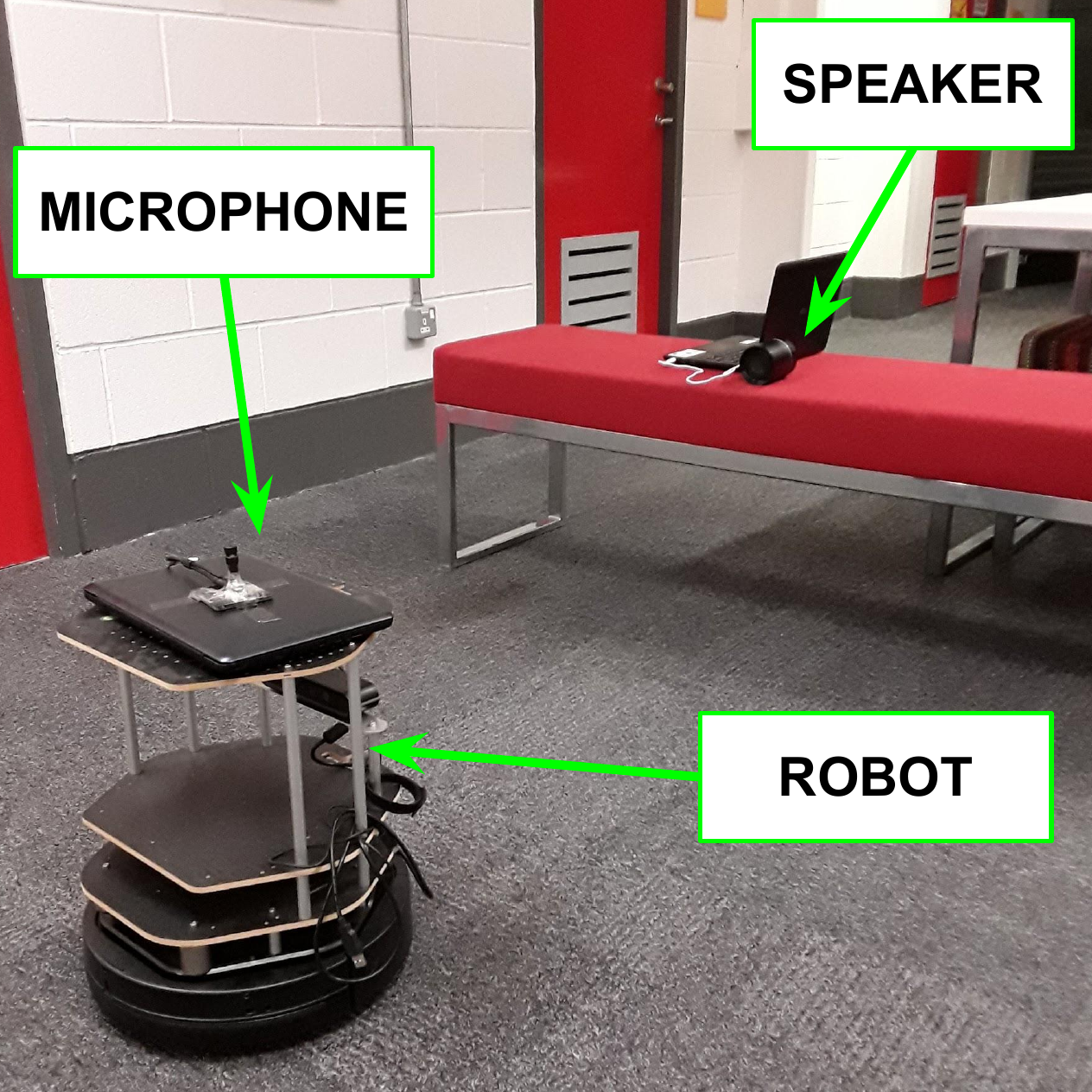}
		\caption{Experiment setup for collecting sound data}
		\label{fig_sound_experiment}
	\end{subfigure}\hfill
	\begin{subfigure}[b]{0.4\textwidth}
		\centering
		\includegraphics[width=\textwidth]{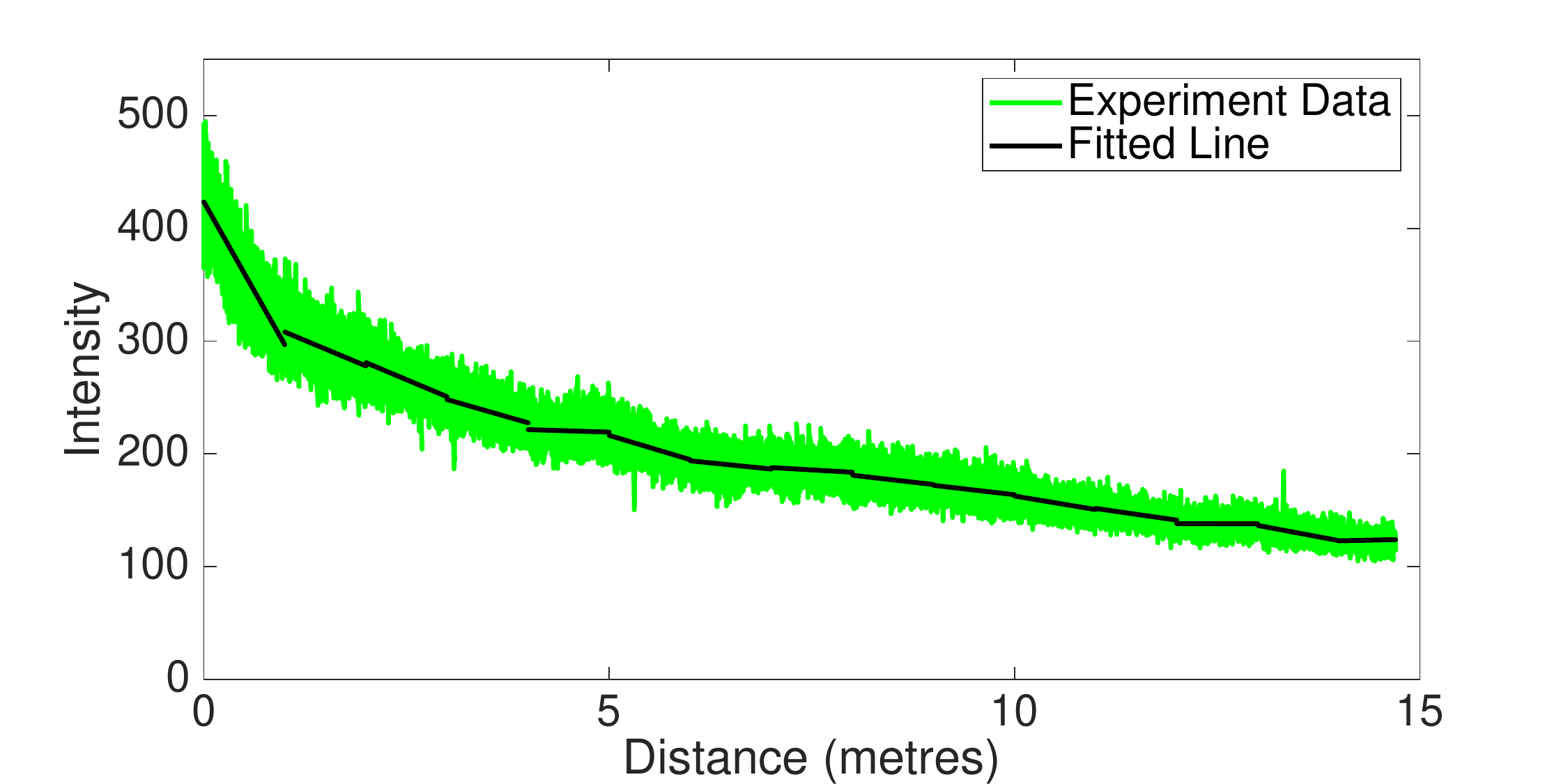}
		\caption{Modelling sound degradation with distance\\\hfill}
		\label{fig_line-fits-noise-model}
	\end{subfigure}\hfill
	\begin{subfigure}[b]{0.4\textwidth}
		\centering
		\includegraphics[width=\textwidth]{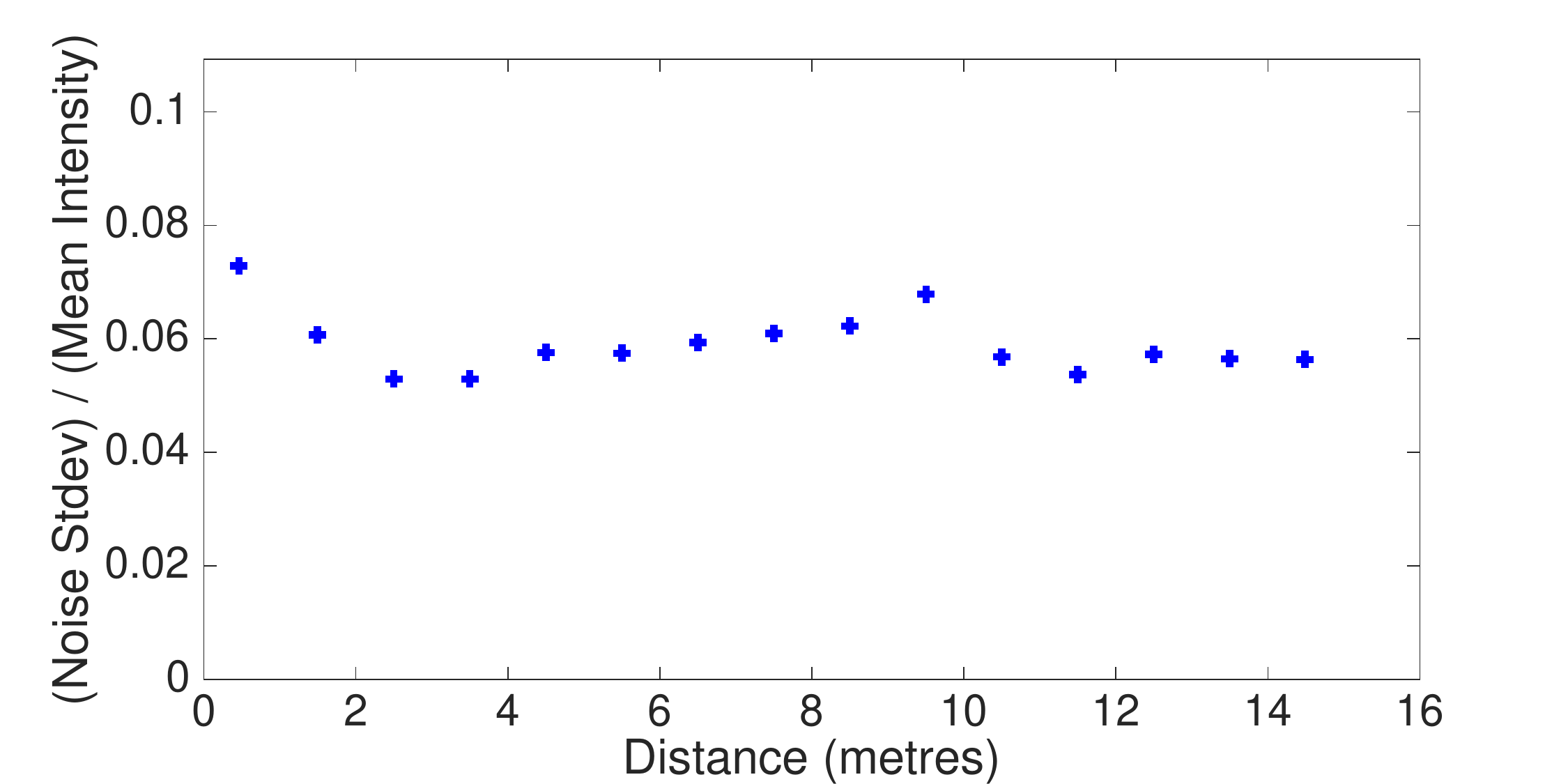}
		\caption{Ratio of deviation to mean intensity within each 1-metre segment of line fit}
		\label{fig_noise-stdev-to-mean-intensity}
	\end{subfigure}
	\caption{Modelling the degradation of sound with distance\label{fig_sound_modelling}}
\end{figure*}
\section{Swarm Communication Experiments and Results (Realistic Communication)}\label{sec_sound_modelling_results}
\subsection{Communication Modelling}\label{subsec_sound_modelling}

The successful deployment of the Rep-Att algorithm on real robots hinges on development of suitable communication hardware, a good visual system for detecting targets and a number of navigation, transport and manipulation challenges. We show in this section our progress in realising the communication mechanism for the robots on hardware. We show that gradient computation can be realized on real robot platforms using sound, and that the algorithm continues to work well with a realistic, noisy sound model. The first step involves collection of sound data using a robot equipped with an omnidirectional microphone to measure sound intensity from a sound source (Fig. \ref{fig_sound_experiment}). In our experiment, the robot started 15m away from a speaker broadcasting white noise signal and was made to move toward the speaker at a speed of 0.1m/s. The robot logged distance travelled (using odometry) and sound intensity as it journeyed toward the speaker. This experiment was repeated five times in order to collect reliable data for developing the model equation.

In the second step, the sound data was used to model its degradation as distance of propagation increases, represented as Equation \ref{eqn_sound_model} \cite{Yu2017}. The $A_{e}$ term was added to account for the mean ambient noise in the environment. $A_{0}$ is the sound intensity from the source, $\alpha$ is the attenuation factor of the environment and $d$ is the distance between the sound source and microphone. The parameters of Equation \ref{eqn_sound_model} were computed by finding the least square fit between the collected data and the model equation using a simple MATLAB script. The computed parameters were $A_{0} = 140.5193$, $\alpha = 0.1193$  and $A_{e} = 48.1824$.

\begin{equation}
 A^{k}_{ij} = A_{0}e^{-\alpha d_{ij}} + A_{e}
 \label{eqn_sound_model}
\end{equation}

  The third step involved quantifying the noise in the recorded sound signal (shown in Fig. \ref{fig_line-fits-noise-model}). To do this, the logged data were broken into 1m segments; a line was fitted to each segment; and the deviation of recorded data on each segment from the fitted line was computed. We observed a direct proportional relationship between sound intensity and the computed deviation. However, there was minimal variation in the ratio of the deviation to mean intensity for each segment (Fig. \ref{fig_noise-stdev-to-mean-intensity}). Thus, we modelled the noise in the signal as the mean of this ratio (0.06 of sound intensity). The final sound intensity model that takes into consideration the noise is as shown in Equation \ref{eqn_noisy_sound_intensity}, where $A_{pure}= A^{k}_{ij}$ computed using Equation \ref{eqn_sound_model} and $\mathcal{N}(0,0.06)$ is a normal distribution with mean of 0 and deviation of 0.06.
  
  \begin{equation}
	  A_{noisy} = A_{pure}(1-\mathcal{N}(0,0.06))
	  \label{eqn_noisy_sound_intensity}
  \end{equation}
 
 \begin{figure}[t]
 	\centering
 	\includegraphics[width=0.4\textwidth]{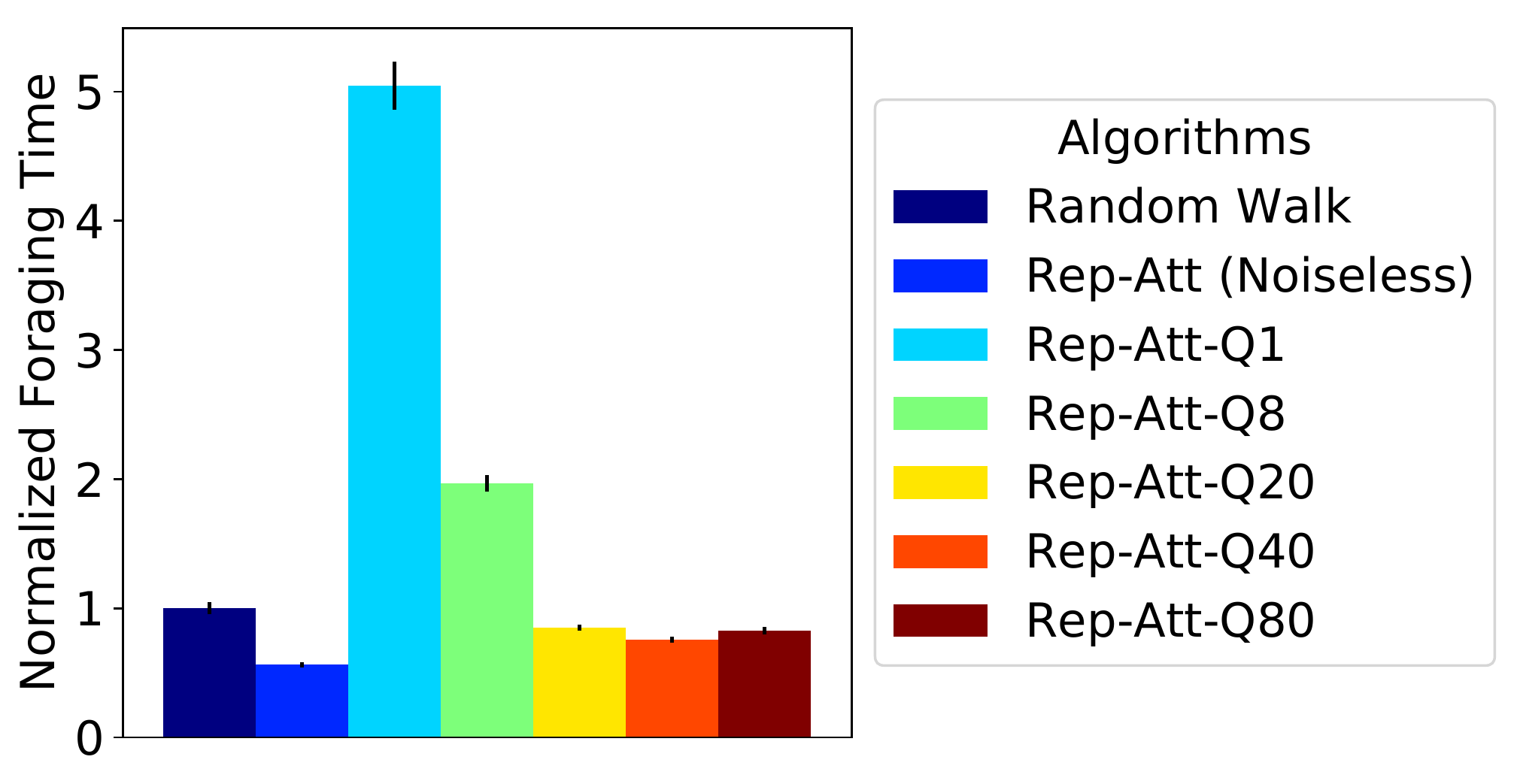}
 	\caption{Investigating the effect of average filter on the performance of Rep-Att using realistic noise values. The filtering window was increased in steps from 1 to 80 to visualize its effect.}
 	\label{fig_2Cluster-RepAt-Noise-v-Q-Updated}
 \end{figure}

  \begin{figure*}[t]
  	\centering
  	\begin{subfigure}[b]{0.6\textwidth}
  		\centering
  		\includegraphics[width=\textwidth]{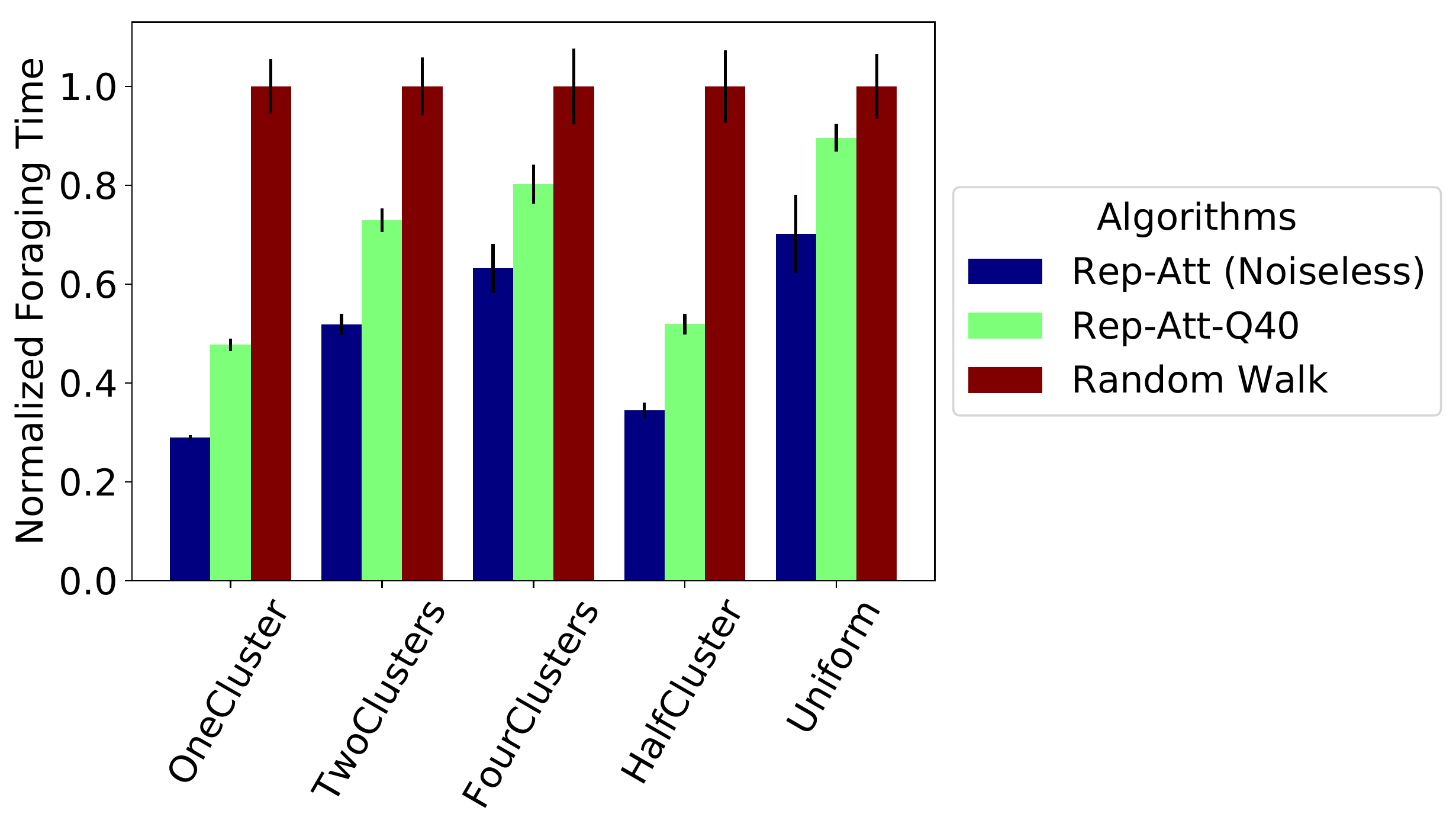}
  		\subcaption{50m x 50m worlds}
  		\label{fig_Normalized-Plot-50m-by-50m-World}
  	\end{subfigure}
  	\hfill
  	\begin{subfigure}[b]{0.39\textwidth}
  		\centering
  		\includegraphics[width=\textwidth]{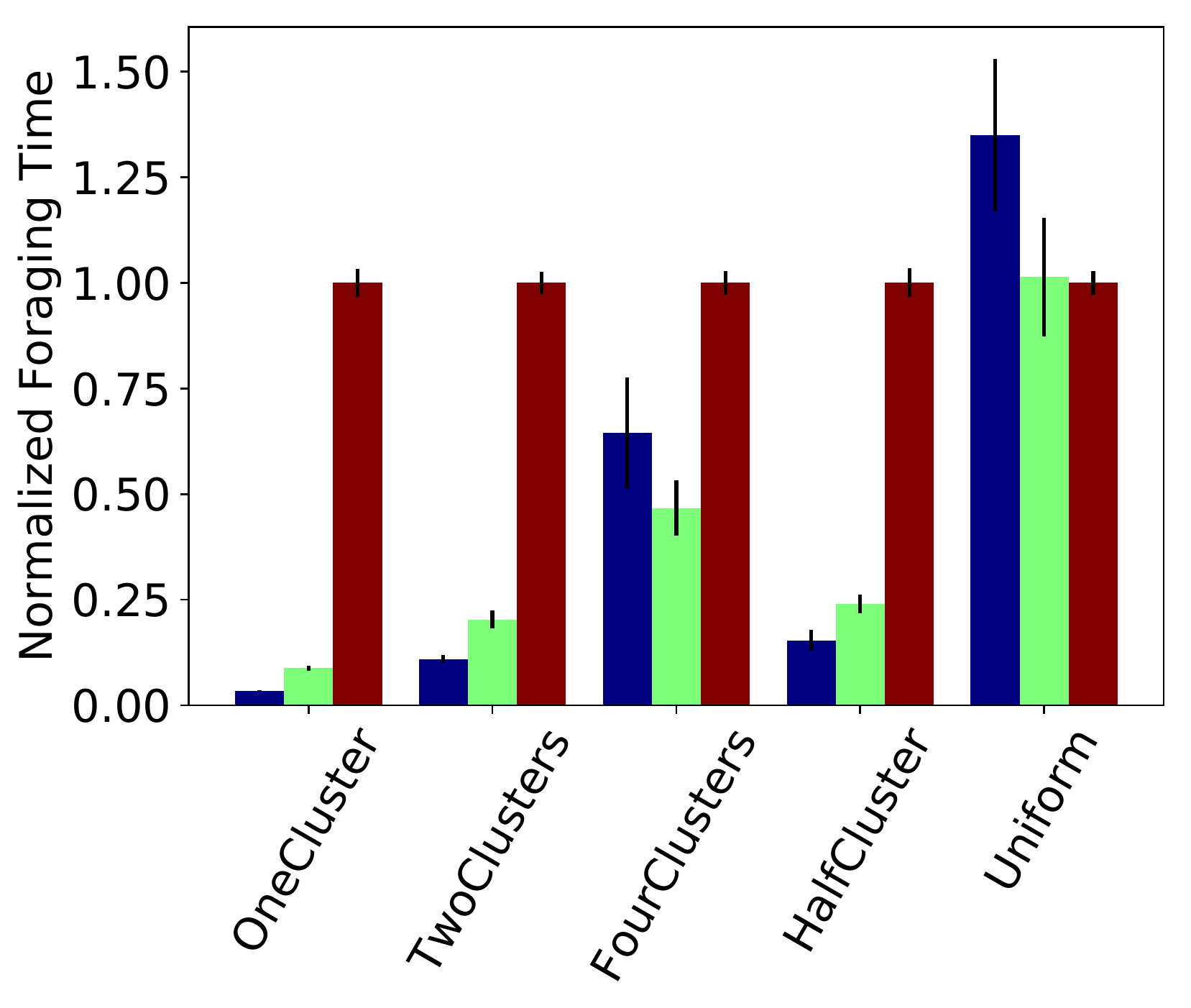}
  		\subcaption{100m x 100m worlds}
  		\label{fig_Normalized-Plot-100m-by-100m-World}
  	\end{subfigure}

  	%
  	\caption{Time taken in seconds to return 90\% of targets to nest in 5 different world scenarios. Each bar represents the mean of 30 simulation repetitions. The error bars represent 95\% confidence interval.}
  	\label{fig_normalized-plots-realistic-sound}
  \end{figure*}
After obtaining this realistic sound model we introduced it to the simulator in place of Equation \ref{eqn_linear_intensity} and repeated the original simulations. This had an extremely detrimental effect as can be seen by comparing Rep-Att-Q1 to Random Walk in Fig. \ref{fig_2Cluster-RepAt-Noise-v-Q-Updated}. Indeed, the Rep-Att algorithm was now five times slower than random walk. We examined the sensory data from these simulations (not shown) and arrived at the following explanation. With a simulation update frequency of 40Hz, the change in the underlying sound intensity from timestep to timestep is very small compared to the magnitude of the noise. As such, the chances that the signal will have increased or decreased is very nearly 50:50 regardless of the underlying signal. Not only does this mask any useful gradient information, but it also leads to a significant increase in overall turn probability because the effect of multiplying and dividing by $G_{a/r}$ is not symmetric in terms of the expected number of turns. 

To mitigate the effect of noise on the Rep-Att algorithm, we introduced an averaging filter to be used by the robots when sensing attraction and repulsion signals. This simple filter computes the average of a queue of multiple instantaneous signal value measurements and the result is used to compute gradient information by comparing it to the mean of the previous queue. The effect of this filter in a foraging scenario is shown in Fig. \ref{fig_2Cluster-RepAt-Noise-v-Q-Updated} for the two clusters target distribution, comparing the random walk, noiseless communication signal and noisy communication signal with queue from 1 to 80. This shows that the filter is effective, and an average of 40 samples gives best foraging performance.
%

The realistic communication model with an averaging filter of queue size 40 was tested on the five target distributions investigated in Section \ref{sec_results}. Figure \ref{fig_Normalized-Plot-50m-by-50m-World} shows the relative foraging performance of the Rep-Att algorithm with noiseless (queue size 1) and noisy (queue size 40) communication, compared to the baseline Random Walk algorithm. The noisy communication resulted in a maximum of 39.30\% decrease in performance when compared with the noiseless communication (this was observed in the one cluster distribution). Notwithstanding, Rep-Att still outperforms the baseline algorithm with a minimum 10.38\% improvement in foraging time (based on comparisons on uniform target distribution), with the maximum improvement being 52.27\% (based on comparisons on one cluster target distribution).

A final test on the algorithm's performance was conducted on a larger world, with 100m x 100m dimension. The target locations remained unchanged from the 50m by 50m worlds, so there was an area of empty space between the targets and the bounding wall. The bar chart in Fig. \ref{fig_Normalized-Plot-100m-by-100m-World} shows the normalized bar plots. It is evident that with increase in world dimensions, Rep-Att's superiority over the baseline increased substantially in all cases except the uniform world. Surprisingly, the version with noisy communication and a queue size of 40 outperforms the noiseless verions with queue size 1 in the four cluster and uniform environments. The relative contributions of the noise and longer queue size to this phenomenon are unclear and will be investigated in future. 

\subsection{Communication Validation Experiment}
\begin{figure}[t]
	\centering
	\includegraphics[height=3.3cm]{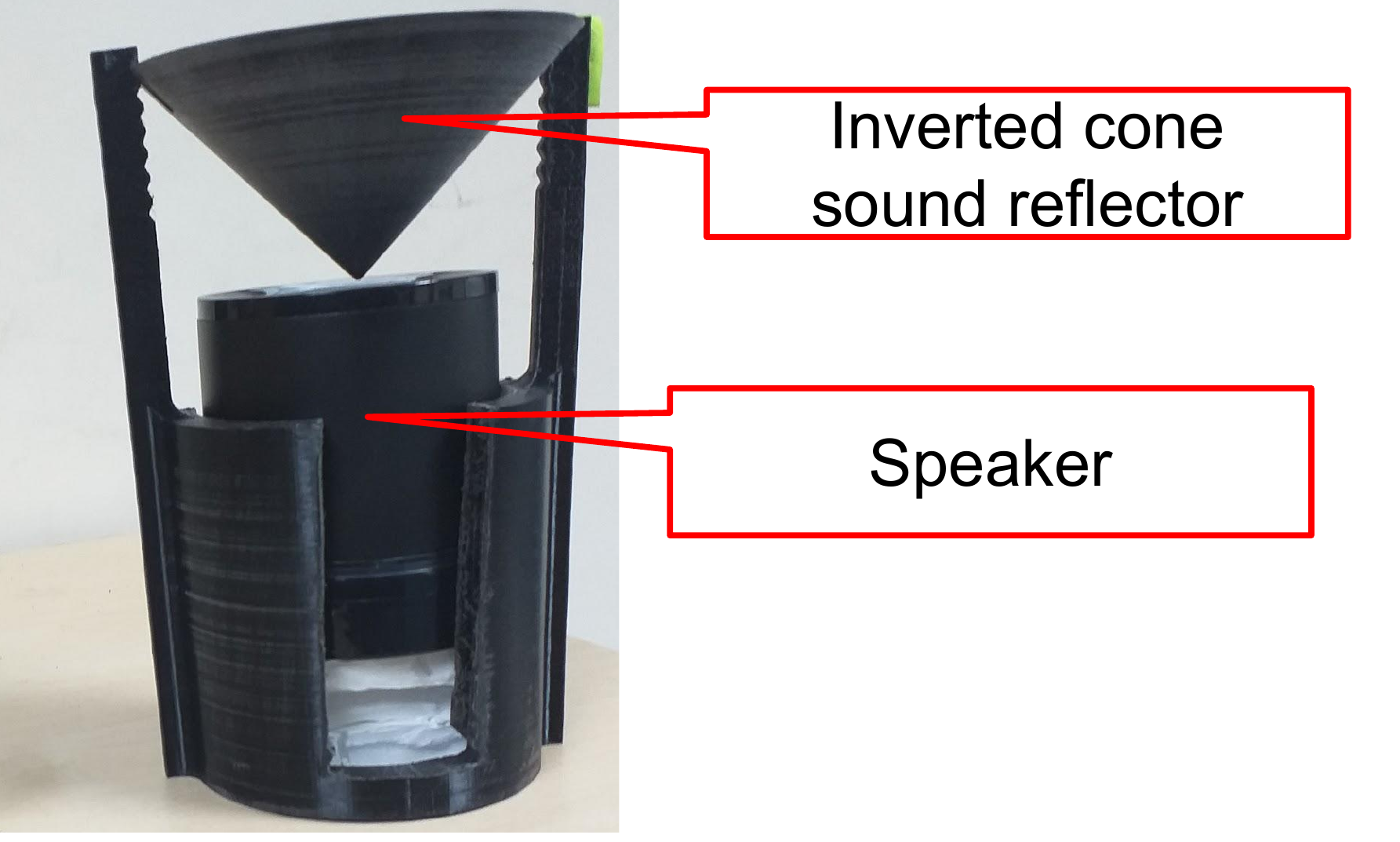}
	\caption{3D-printed setup for makes use of an inverted cone to reflect the sound signals omni-directionally round the speaker.}
	\label{fig_omni-speaker}
\end{figure}

To validate that our communication model is a valid representation of the real hardware we initially characterised, we set up an experiment with a single Turtlebot2 in a 5m x 3m arena. The robot had a omnidirectional microphone and was programmed to use the attraction behaviour from our algorithm to search for a single sound source. The setup in Fig. \ref{fig_omni-speaker} was used to reflect the sound from a speaker in all directions, effectively making it omnidirectional. To the best of our knowledge, this is the first such experiments to be conducted in real hardware using bio-inspired temporal gradient computation from a sound signal. 

\begin{figure*}[t]
	\centering
	\begin{subfigure}[b]{0.23\textwidth}
		\centering
		\includegraphics[width=\textwidth]{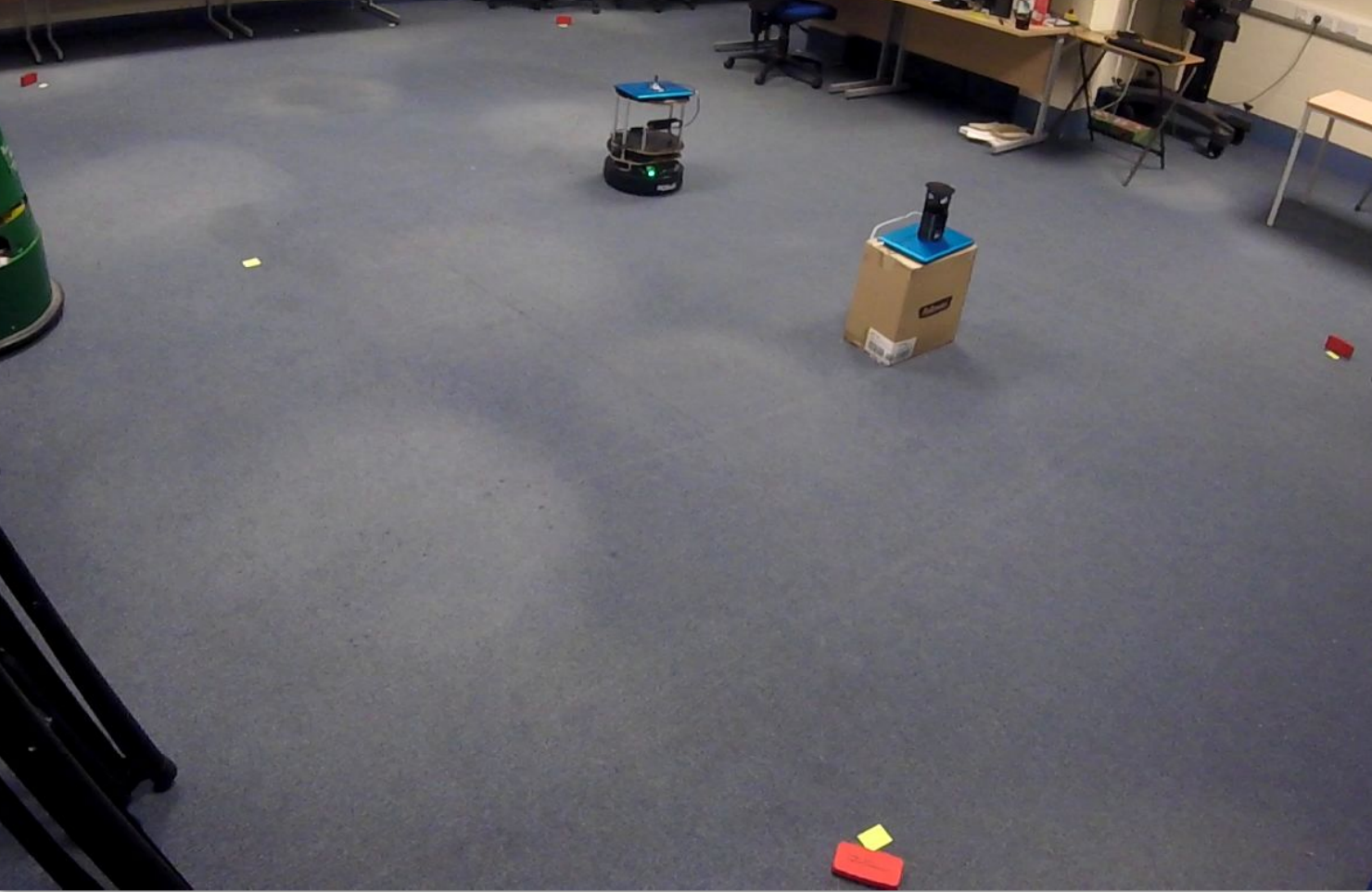}
		\subcaption{time = 0s}
		\label{fig_GOPR9034-0m}
	\end{subfigure}
	\hfill
	\begin{subfigure}[b]{0.23\textwidth}
		\centering
		\includegraphics[width=\textwidth]{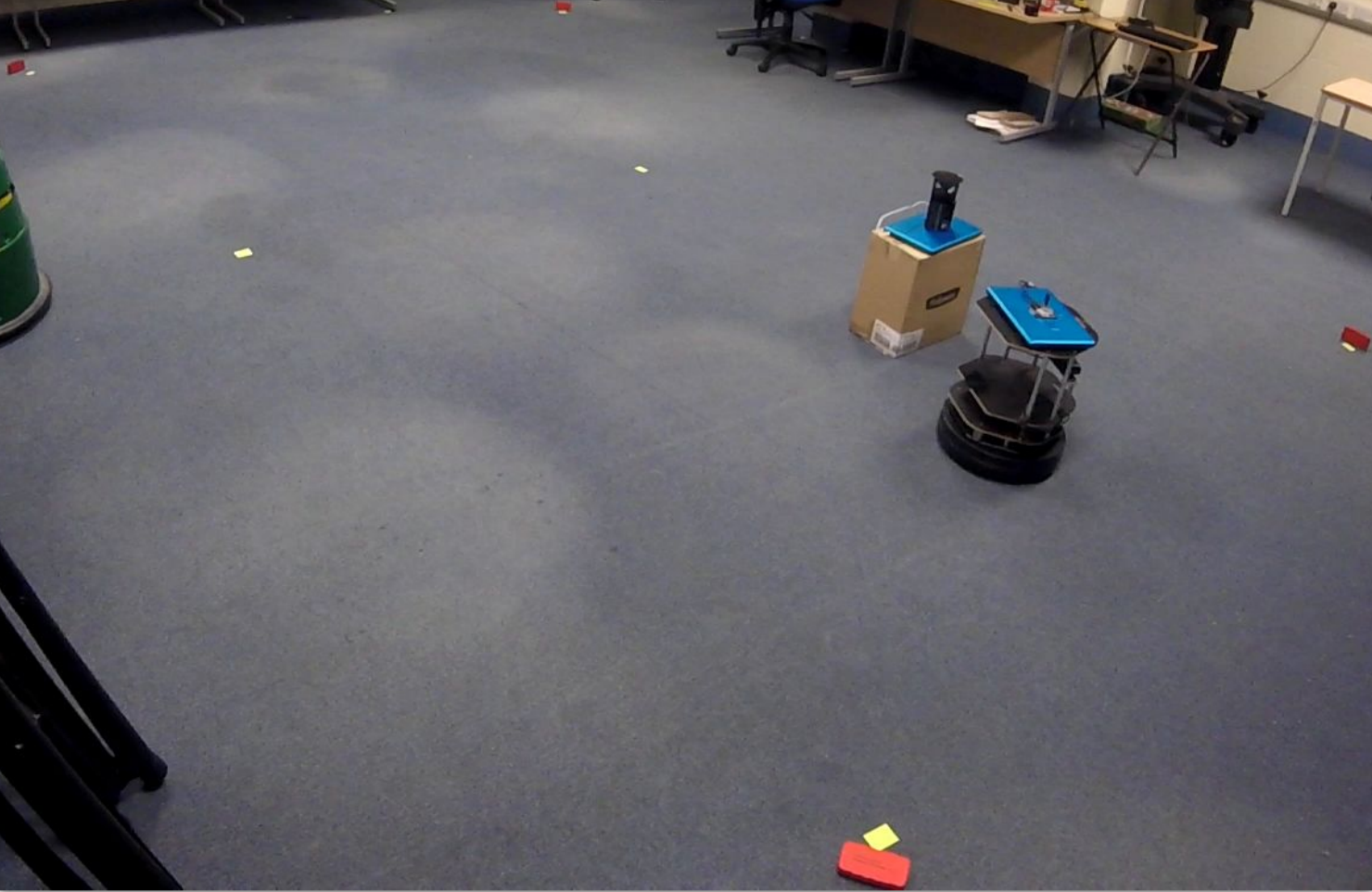}
		\subcaption{time = 480s}
		\label{fig_GOPR9034-8m}
	\end{subfigure}
	\hfill
	\begin{subfigure}[b]{0.23\textwidth}
		\centering
		\includegraphics[width=\textwidth]{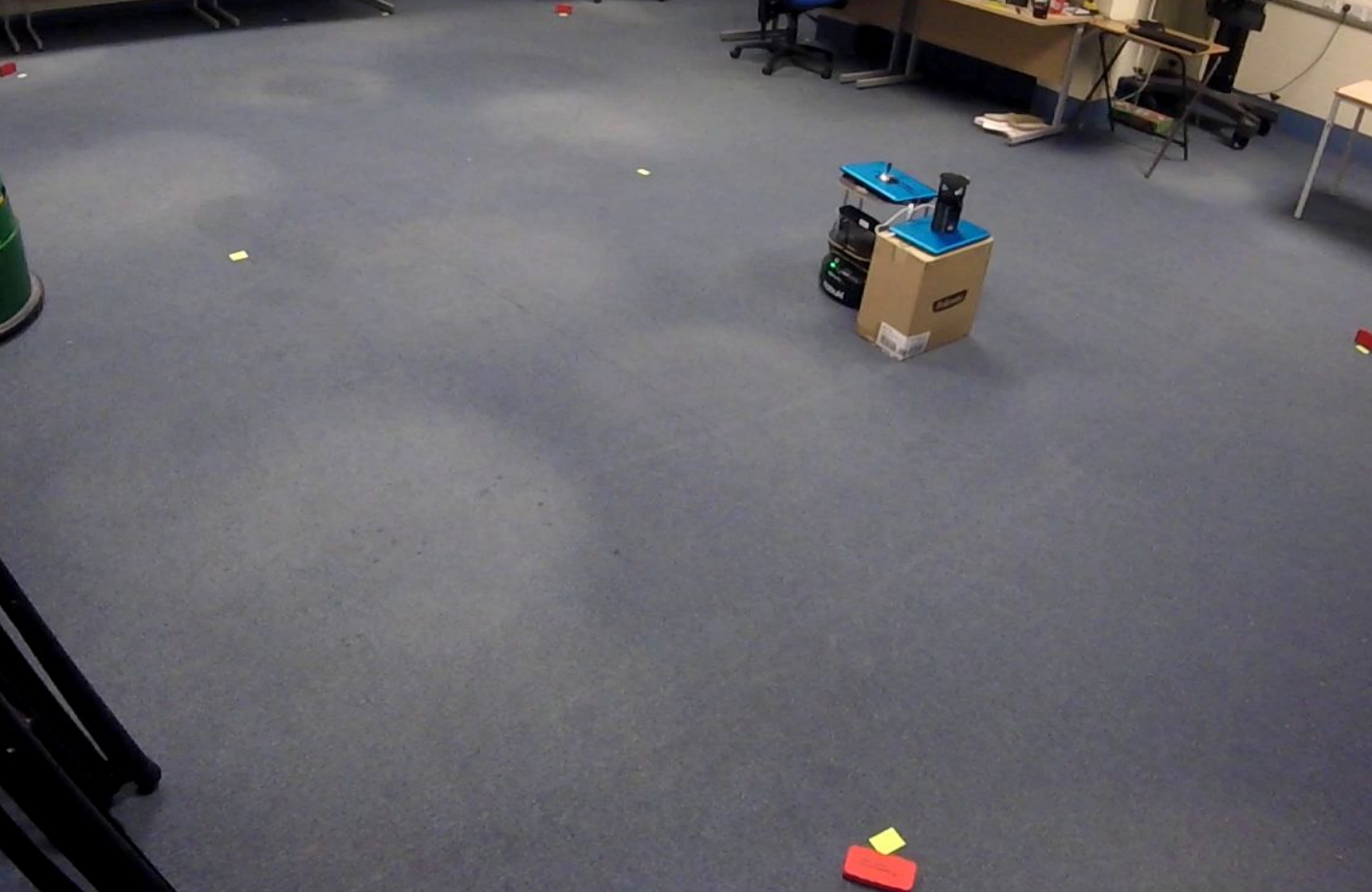}
		\subcaption{time = 720s}
		\label{fig_GP019034-12m}
	\end{subfigure}
	\hfill
	\begin{subfigure}[b]{0.23\textwidth}
		\centering
		\includegraphics[width=\textwidth]{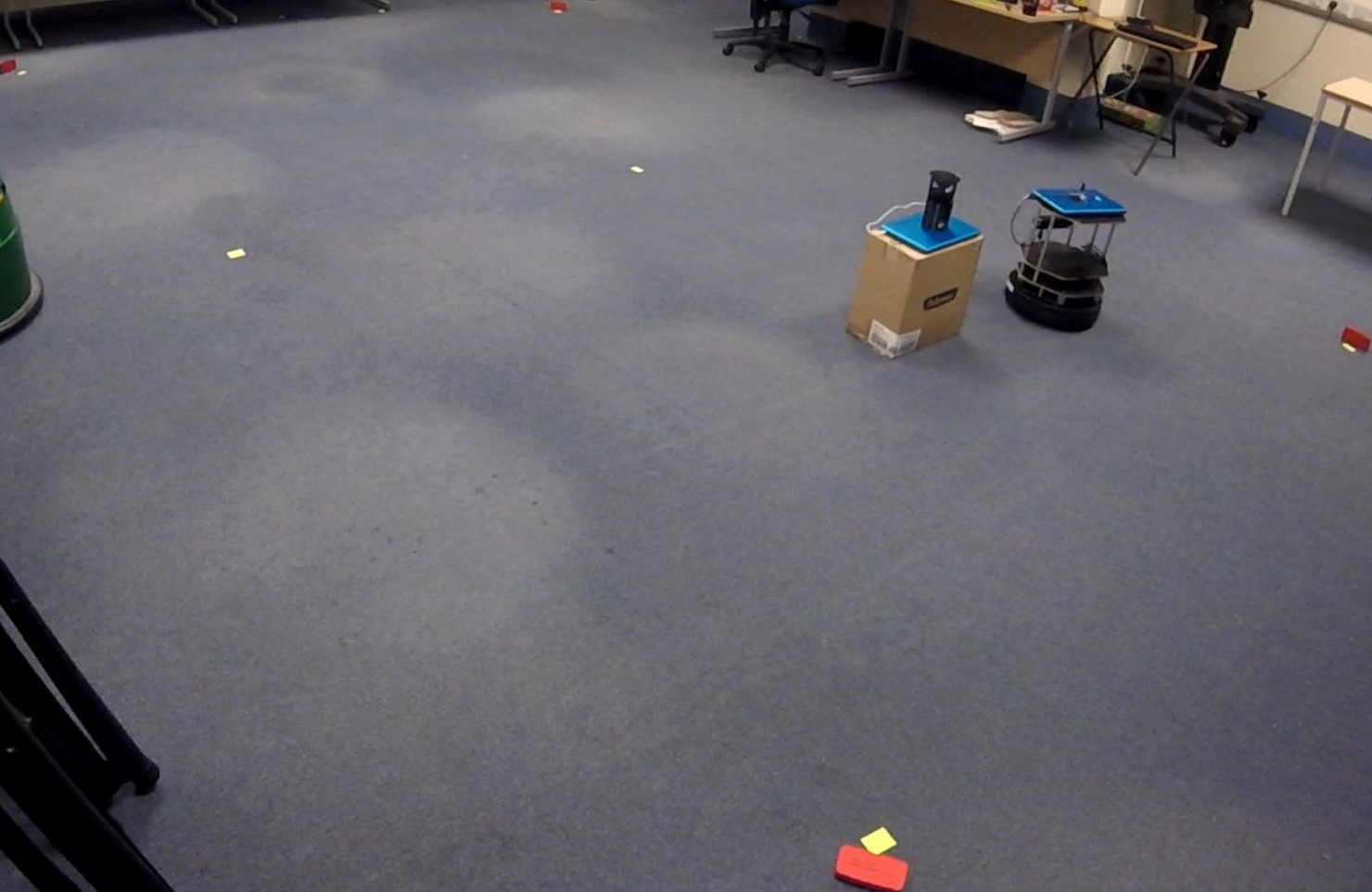}
		\subcaption{time = 960s}
		\label{fig_GP019034-16m}
	\end{subfigure}
	\caption{Experiment with a Turtlebot2 using a single microphone and performing chemotaxis to be attracted to white noise from an omnidirectional speaker. The turtlebot is restricted to search within a 5m by 3m virtual wall (enabled by odometry). Footage of the full experiment is available in supplementary video file.}
	\label{fig_white-noise-attraction}
\end{figure*}

The robot velocity used was 0.1 m/s and a filter queue size of 40 samples was used. Other parameters were left unchanged (Table \ref{table_parameters}). The sound source was kept at (-1.5,0), while the Turtlebot2 starting location was (0,0) facing away from the sound source. The Turtlebot2's odometry was used to keep track of where it is in the arena. Each experiment lasted for 1000 seconds and was repeated 10 times. We investigate our algorithm's ability to find the sound source and provide results for pure random walk, as a baseline. We conducted equivalent experiments in simulation and using the real hardware. 

Fig. \ref{fig_white-noise-attraction} shows snapshots of the experiment, where the Turtlebot2, when controlled with Rep-Att, was able to find the sound source and remain within that region for the remainder of the experiment's duration. 

\begin{figure}[t]
	\centering
	\includegraphics[width=0.5\textwidth]{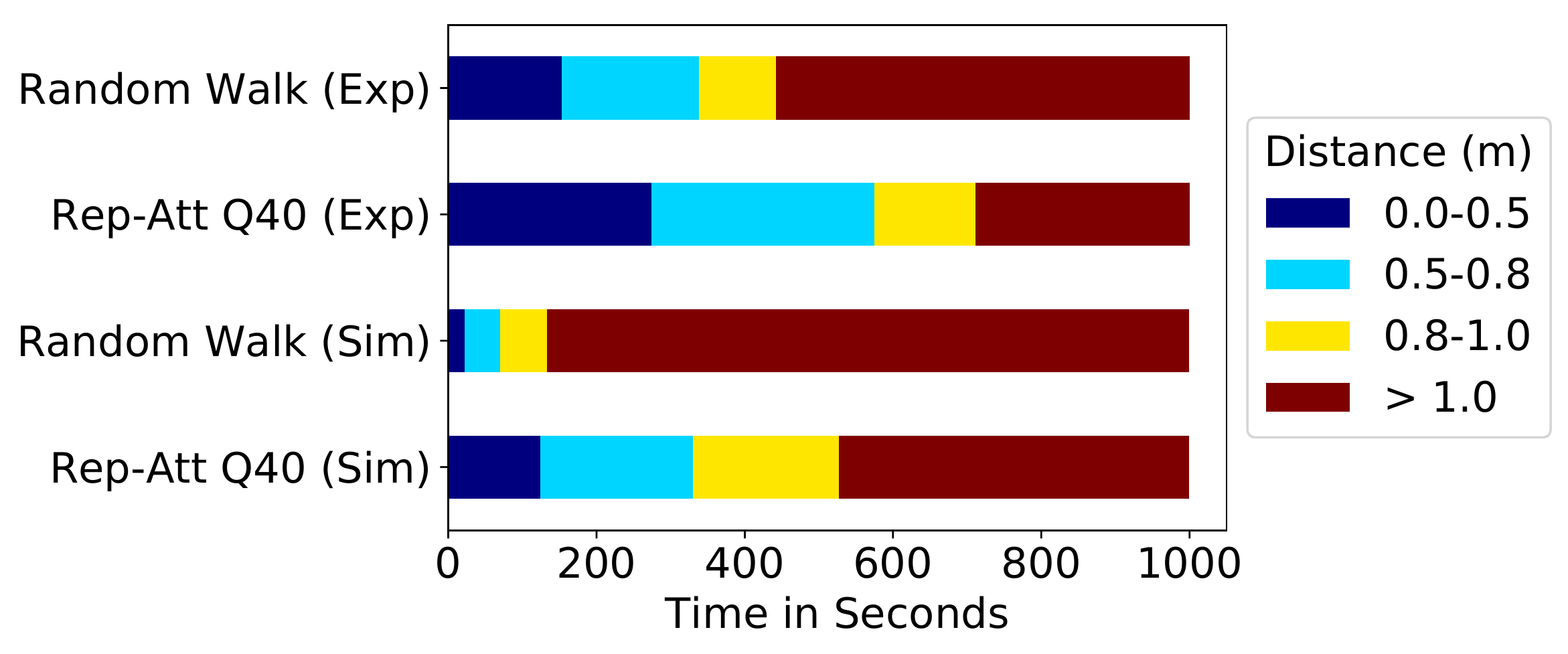}
	\caption{Proportion of 1000 seconds experiment time spent by Turtlebot2 relative to the sound source. Please note that robot bumps into the sound source setup when it is within at least 0.27m away from it. \textit{Exp} represent data gotten from Turtlebot2 experiment, \textit{Sim} represents data gotten from equivalent Gazebo simulation, \textit{Rep-Att Q40} represents noisy communication with window/queue size of 40 samples.}
	\label{fig_sound-source-localization-stacked}
\end{figure}

In Quantitative terms, Fig. \ref{fig_sound-source-localization-stacked} shows the proportion of the experiment time a robot spent in regions around the sound source. Each real robot experiment was repeated 10 times, while simulation experiments were repeated 30 times. The results show that both real and simulated robots effectively navigated to the sound source when using the taxis (Rep-Att) behaviour and spent significantly more time within 1m of the sound source in comparison to random walk. The relatively good agreement between simulated and physical experiments suggests that the sound model we used is a meaningful reflection of real hardware. The noticeable difference between simulation and hardware experiment times for both algorithms is due to slight differences between the two platform implementations. For example, in the hardware experiment, the Turtlebot2's obstacle avoidance behaviour entailed its reversing for 10cm before turning a random angle, which meant it spent more time within the obstacle's region (wall or sound source) when compared to the time taken by the simulated robot (see obstacle avoidance description in Section \ref{sec_rep_att})

\section{Conclusion}\label{sec_conclusion}
We have presented Rep-Att, a simple yet effective, swarm foraging algorithm that works surprisingly well - even when given a realistic and far from ideal communication model strongly grounded in real hardware experiment. Rep-Att is based on a concept of selective broadcast of repulsion and attraction signals among swarm agents, which was used to adapt robot's turning probability while searching. The end result was a significant improvement in the swarm's foraging efficiency. In addition to extensive simulation studies on how our algorithm performs under different target distributions, we have also validated our communication model by comparing hardware to simulated results for a single robot.

Future work will involve a full demonstration of our foraging algorithm on a swarm of real robots. We also intend to extend the capability of our robots to use odometry in conjunction with memory of previous places targets were last found so that they can weigh the options between returning to last foraged locations or solely search using repulsion-attraction signals. Investigation of the effects of obstacles within the foraging environment will also be looked into.

\IEEEpeerreviewmaketitle

\bibliographystyle{plainnat}
\bibliography{./RSS19_JB}

\end{document}